\documentclass[review]{elsarticle}

\usepackage{lineno,hyperref}
\modulolinenumbers[5]

\journal{Journal of \LaTeX\ Templates}

\usepackage{hyperref}
\usepackage[utf8]{inputenc} 
\usepackage{graphicx}
\usepackage{bm}
\usepackage{mathrsfs}
\usepackage{amsmath}
\usepackage{amssymb}
\usepackage{algorithm}
\usepackage{algorithmic}
\usepackage{array}
\usepackage{makecell}
\usepackage{booktabs}
\usepackage{caption3}
\usepackage{textcomp} 
\usepackage{gensymb} 
\usepackage{url}
\usepackage{multirow}
\usepackage{xcolor}
\newcommand{\eg}{e.g.}
\newcommand{\ie}{i.e.}
\newcommand{\etal}{\textit{et al.~}}

\begin{document}

\begin{frontmatter}

\title{Collaborative representation-based robust face recognition by discriminative low-rank representation}

\author[add1,add2]{Wen Zhao}

\author[add1,add2]{Xiao-Jun Wu\corref{mycorrespondingauthor}}
\cortext[mycorrespondingauthor]{Corresponding author}
\ead{wu\_xiaojun@jiangnan.edu.cn}

\author[add1,add2]{He-Feng Yin}

\author[add1,add2]{Zi-Qi Li}

\address[add1]{School of Internet of Things Engineering, Jiangnan University, Wuxi 214122, China}
\address[add2]{Jiangsu Provincial Laboratory of Pattern Recognition and Computational Intelligence, Jiangnan University, Wuxi 214122, China}

\begin{abstract}
We consider the problem of robust face recognition in which both the training and test samples might be corrupted because of disguise and occlusion. Performance of conventional subspace learning methods and recently proposed sparse representation based classification (SRC) might be degraded when corrupted training samples are provided. In addition, sparsity based approaches are time-consuming due to the sparsity constraint. To alleviate the aforementioned problems to some extent, in this paper, we propose a discriminative low-rank representation method for collaborative representation-based (DLRR-CR) robust face recognition. DLRR-CR not only obtains a clean dictionary, it further forces the sub-dictionaries for distinct classes to be as independent as possible by introducing a structural incoherence regularization term. Simultaneously, a low-rank projection matrix can be learned to remove the possible corruptions in the testing samples. Collaborative representation based classification (CRC) method is exploited in our proposed method which has closed-form solution. Experimental results obtained on public face databases verify the effectiveness and robustness of our method.
\end{abstract}


\end{frontmatter}


\section{Introduction}
\label{sec-1}
Face recognition (FR), as one of the biometric technologies, has been raised concern due to its critical applications in many real-world scenarios, including access control, social networks, card identification, digital entertainment and intelligent interaction~\cite{ding2015robust},~\cite{zhen2015muscular}. So the improvement of FR technology has meaningful value in developing a modern city. It is well-known that the original face images usually have high dimensionality, leading to high computational complexity. Due to the fact that face images usually reside on a low dimensional subspace, many feature extraction techniques, such as Eigenfaces~\cite{turk1991eigenfaces}, Fisherfaces~\cite{belhumeur1997eigenfaces}, Laplacianfaces~\cite{he2005face} and their variants~\cite{xiao2004new,zheng2006nearest,zheng2006reformative}, have been developed. Then a corresponding classifier such as the nearest neighbor classifier (NNC) and support vector machine (SVM) can be used. When face images are taken under a well controlled setting, the above subspace learning methods can shown promising results on FR under the circumstances. Nevertheless, usually robust FR is a challenging issue due to the appearance of variability in variance, illumination, pose, occlusion and even disguise. Moreover, from another point of view, the effectiveness of the learned subspace will be degraded due to small amount of training samples. Luo \etal\cite{luo2015image} proposed a selective regularized subspace learning (SRSL) algorithm which learns a local subspace for each sample instead of learning a global subspace for all samples.

In recent years, sparse representation-based classification (SRC)~\cite{wright2008robust} as a new robust FR framework has been proposed, which represents each test image as a sparse linear combination of the whole training images. Then by solving an $\ell_1$-minimization optimization problem~\cite{donoho2006most}, the results of the identification can be achieved by the minimum class-specific reconstruction error. If the test image is corrupted due to occlusion or corruption, SRC is able to show good robustness by introducing an identity matrix as the occlusion dictionary. However, because of the high dimensionality of the identity matrix, the sparse coding procedure would be computationally expensive. An extended SRC (ESRC) method was proposed by Deng \etal \cite{deng2012extended} in order to solve the above issue, which uses an auxiliary intra-class variant dictionary by subtracting the neutral image from the other images of the same subject to represent the possible corruption. The variant matrix with much smaller dimensionality is then used as occlusion dictionary appended to the raw training data. However, the auxiliary intra-class variant dictionary might not depict the various corruptions of face images accurately. Additionally, Zhang \etal \cite{yang2013gabor} proposed Gabor feature based SRC (GSRC) method, in which the image Gabor features are used for SRC. The use of Gabor kernels makes the occlusion dictionary compressible, which reduces the computational cost to some extent. Although both ESRC~\cite{deng2012extended} and GSRC~\cite{yang2010gabor} are more robust to deal with test images with occlusion or corruption than SRC, it is still time-consuming to solve the $\ell_1$-minimization optimization problem.

What is more, very recently some works \cite{rigamonti2011sparse}, \cite{zhang2011sparse}, \cite{wang2010locality}, \cite{shi2011face} have started to show doubt about the necessity of $\ell_1$-norm-based sparsity, such as described in \cite{wright2008robust}, to the improvement of the performance in face image classification. Meanwhile, the sparsity based classification schemes such as SRC are very computationally expensive \cite{yang2010gabor}. Zhang \etal \cite{zhang2011sparse} replaced the $\ell_1$-norm by the $\ell_2$-norm to emphasize the role of collaborative representation (CR), which represents the test image collaboratively by using the whole training images from all classes. The above problem has been examined by some recent works directly or indirectly. One of the example is Rigamonti \etal proposed in \cite{rigamonti2011sparse}, they used two different data models, the first one is the sparse representation based on the $\ell_1$-norm and the second one is proposed by passing the input signal into a simple convolution filter, to compared the discrimination of themselves. As a result, a similar recognition accuracy could be achieved by these two models. Therefore, some scholars drew a conclusion that $\ell_1$-norm-based sparsity is not as important as it claims in the previous methods. Under this awareness, by replacing the $\ell_1$-norm with the $\ell_2$-norm, a very simple yet much more valid face classification scheme was proposed by Zhang \etal \cite{zhang2011sparse} named CRC. The collaborative representation technique has been adopted in many papers \cite{zhang2015ccr}, \cite{chi2013classification} in recent years.

The role of collaboration between different subjects in linear representation of the test image is considered by CRC, it employs the much weaker $\ell_2$-norm instead of the $\ell_1$-norm to regularize the coefficient representation and has very competitive FR accuracy but significantly lower complexity. It is unfortunate that when the training and test samples are corrupted simultaneously \cite{chen2012low}, the effectiveness of CRC would be dropped. To alleviate the aforementioned problems \cite{de2003framework}, \cite{candes2011robust}, many recent works on low-rank matrix recovery (LRMR) have been proposed. Among these works, the robust PCA (RPCA) method presented by Wright \etal \cite{wright2009robust} assumed that all data reside on a single subspace, and then a low-rank data matrix can be recovered by separating occluded pixels and various corruptions from the original training samples. Actually, the face image samples usually come from multiple subspaces, which will affect the performance of RPCA. A new method was proposed by Liu \etal \cite{liu2010robust}, \cite{liu2012robust} named low-rank representation (LRR) under the assumption that data samples are drawn from multiple subspaces. Although LRR can work well to remove the corruptions from the training data, the local structure of data was neglected which might lead to the degradation of recovery performance. There are also some other works proposed in recent years to improve the performance of LRMR, \eg, under the Bayesian framework, Zhao \etal \cite{zhao2014robust} presented a generative RPCA model, the noise in the face images is modeled as a mixture of Gaussians to suit a huge range. Yin \etal\cite{yin2016face} presented a new method called low rank matrix recovery with structural incoherence and low rank projection (LRSI\_LRP) which can correct the corrupted test images with a low rank projection matrix. Zhao \etal\cite{zhao2015collaborative} developed a discriminative low-rank representation method for collaborative representation-based (DLRR-CR) robust face recognition. Chen \etal\cite{chen2018robust} proposed a robust low-rank recovery algorithm (RLRR) with a distance-measure structure for face recognition. Zhang \etal \cite{zhang2014image} proposed a new image classification scheme, which utilizes the non-negative sparse coding, low-rank and sparse matrix decomposition techniques to obtain a classification framework. Zhang \etal \cite{zhang2012mining} presented a new image decomposition technique for an ensemble of associated image samples. This method utilizes a procedure to decompose every face image into three parts: a common image, a low-rank image and a sparse corruption matrix. However, if the test images are corrupted, the performance of the above methods will degrade, because they cannot correct the corrupted test images.

\begin{figure}[h]
\centering
\includegraphics[trim={0mm 0mm 0mm 0mm},clip, width = .8\textwidth]{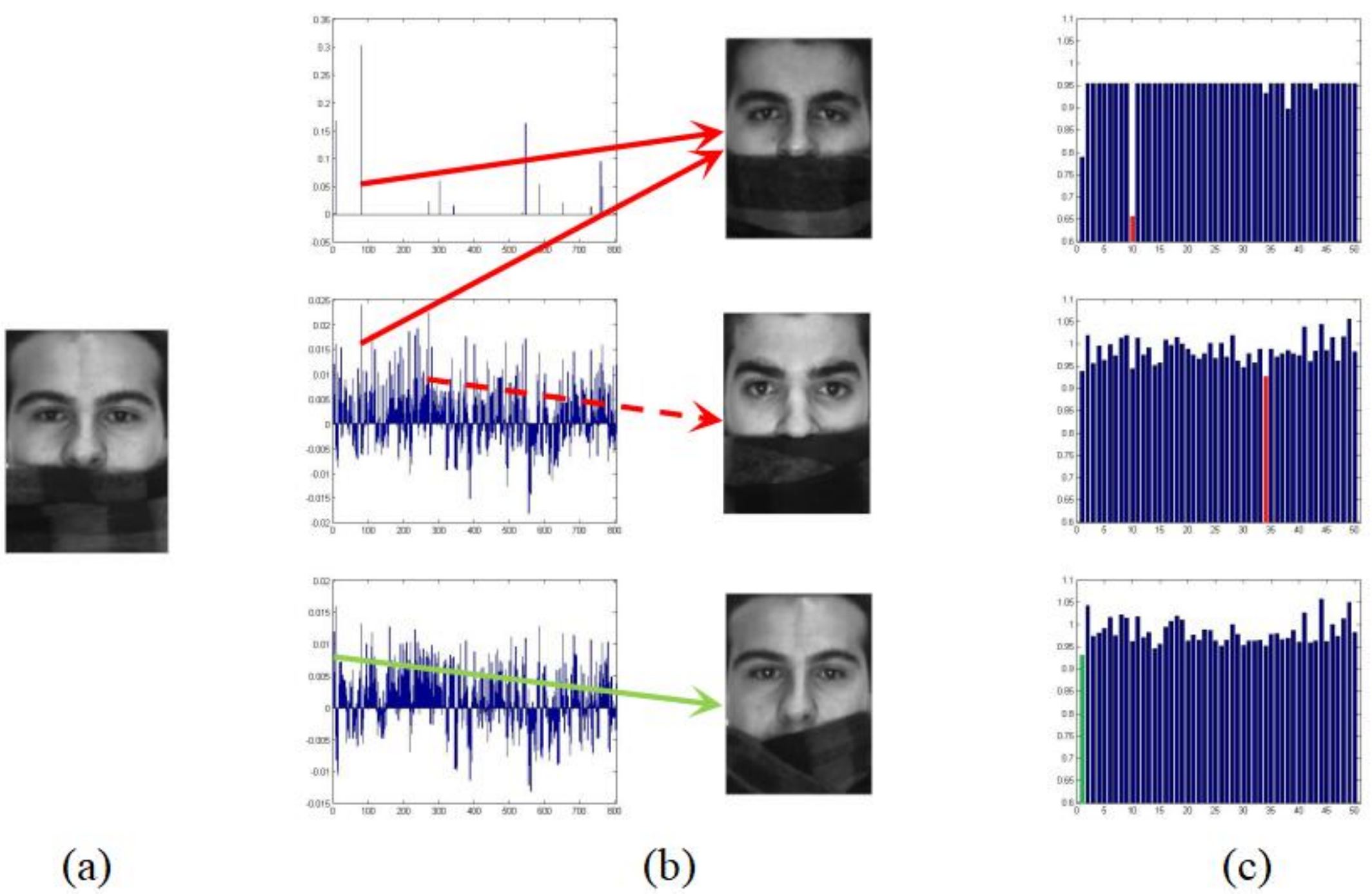}
\caption{Comparison between SRC, CRC and our approach. (a) An original test image occluded by a scarf from the AR database. (b) The first column includes the coefficients of the three methods for the same test image respectively. The second column shows the most similar training images chosen by the three methods. (c) The results of class-wise reconstruction errors of the three methods. Our approach can classify the test image to the correct class while SRC and CRC classify the test image as that class with the most similar training images.}
\label{fig:pro_illus}
\end{figure}

In this paper, to address the problem of robust face recognition, in which both training and test samples might be corrupted by the unknown type, we propose a discriminative low-rank representation method for collaborative representation based (DLRR-CR) method. As revealed by other literatures, if the original training images with corruption are directly used as the dictionary, it will degenerate the performance of FR. To avoid this problem, the proposed method first constructs a discriminative LRR framework to separate the corruptions and recover the clean training samples. The LRR method presented by Liu \etal \cite{liu2012robust} only imposes the low-rank constraint on the representation coefficient matrix of the training samples. In order to reflect incoherence between different classes, a regularization term is added to the formulation of LRR, which can provide additional discriminating ability to our framework and obtain better representations. In addition, a low-rank projection matrix can be learned and then applied to project the corrupted test images into their corresponding underlying subspace to get the corrected test images. It is more importantly to see that our proposed approach is not just used for reconstructing the test samples but for recognition, as illustrated in Fig.~\ref{fig:pro_illus}, the standard SRC and CRC classify the test image as the class with most similar images, while our proposed method alleviates this problem. Obviously, the results of class-wise reconstruction errors of the three methods, shown in Fig.~\ref{fig:pro_illus} (c), exhibit that by using our method the correct class has the smallest reconstruction error, which can demonstrate that our method is more robust to occlusions presented in both training and test images. Section~\ref{sec-3} will present more details. Furthermore, it is worthy to note that in the testing stage, CRC is exploited, which has the observably lower complexity but excellent FR performance. In the experimental part, our method DLRR-CR will show the effectiveness and robustness for FR.

The remainder of this paper is organized as follows. Section \ref{sec-2} briefly reviews some related works on CRC and LRR for FR. In Section \ref{sec-3}, we present the proposed model for FR in detail. Experimental results on real-world face image data are presented in Section \ref{sec-4}. Finally, Section \ref{sec-5} concludes this paper.

\section{Related work}
\label{sec-2}
\subsection{Collaborative representation-based classification (CRC)}
\label{sec-2.1}
We consider $n$ original bases $\mathbf{X}=\left[\boldsymbol{x}_{1}, \boldsymbol{x}_{2}, \ldots, \boldsymbol{x}_{n}\right] \in \mathbb{R}^{m \times n}$ collected from $N$ different classes, and $m$ is the dimension of each base. Class $i$ includes $n_i$ training images which denoted by $\mathbf{X}_i$. $\mathbf{X}$ can be rewritten as $\mathbf{X}=\left[\mathbf{X}_{1}, \mathbf{X}_{2}, \ldots, \mathbf{X}_{N}\right]$. When comes a new test sample $\boldsymbol{y} \in \mathbb{R}^{m}$, SRC aims to find a sparse linear representation coefficient vector $\boldsymbol{\alpha} \in \mathbb{R}^{n}$ so that $\boldsymbol{y}$ can be represented as $\boldsymbol{y}=\mathbf{X} \boldsymbol{\alpha}$. This approximation problem can be calculated via minimizing the following problem:
\begin{equation}\label{equ-1}
\hat{\boldsymbol{\alpha}}=\arg \min _{\boldsymbol{\alpha}}\left\{\|\boldsymbol{y}-\mathbf{X} \boldsymbol{\alpha}\|_{2}^2+\lambda\|\boldsymbol{\alpha}\|_{1}\right\}
\end{equation}
where $\lambda$ denotes a scalar constant, $\|\cdot\|_{2}$ denotes the $\ell_2$-norm, and $\|\cdot\|_{1}$ denotes the $\ell_1$-norm. Many algorithms, such as basis pursuit \cite{chen2001atomic} and Homotopy \cite{yang2011alternating} can be used to figure out the above $\ell_1$-norm minimization problem. The test sample $\boldsymbol{y}$ should lie in the space spanned by the training samples from the correct class. Once we get the solution $\hat{\boldsymbol{\alpha}}$ of (\ref{equ-1}), where $\hat{\boldsymbol{\alpha}}=\left[\boldsymbol{\alpha}_{1} ; \boldsymbol{\alpha}_{2}, \ldots ; \boldsymbol{\alpha}_{N}\right]$ and $\boldsymbol{\alpha}_{i}$ is the representation vector of $\hat{\boldsymbol{\alpha}}$ associated with class $i$, the test sample $\boldsymbol{y}$ can be recognized by the reconstruction error of each class, \ie,
\begin{equation}\label{equ-2}
\textrm{identity}(\boldsymbol{y})=\arg \min _{i}\left\{\left\|\boldsymbol{y}-\mathbf{X}_{i} \hat{\boldsymbol{\alpha}}_{i}\right\|_{2}\right\}
\end{equation}

Based on the fact that face images from different subjects may have similar appearances, thus samples from the uncorrelated classes can participate in representing the test sample $\boldsymbol{y}$. A regularized least square method is developed with significantly lower complexity named CRC, which is formulated as,
\begin{equation}\label{equ-3}
\hat{\boldsymbol{\rho}}=\arg \min _{\boldsymbol{\rho}}\left\{\|\boldsymbol{y}-\mathbf{X} \boldsymbol{\rho}\|_{2}^{2}+\lambda\|\boldsymbol{\rho}\|_{2}^{2}\right\}
\end{equation}
where $\lambda$ is a balance factor. CRC can offer improvements in decreasing computational complexity by using the $\ell_2$-norm-based model. A closed-form solution $\hat{\boldsymbol{\rho}}=\left(\mathbf{X}^{T} \mathbf{X}+\lambda \mathbf{I}\right)^{-1} \mathbf{X}^{T} \boldsymbol{y}$ can be derived by solving (\ref{equ-3}), in which $\left(\mathbf{X}^{T} \mathbf{X}+\lambda \mathbf{I}\right)^{-1} \mathbf{X}^{T}$ can be pre-calculated which leads to the fast computation speed of CRC. In the classification stage, the regularized residuals $\boldsymbol{e}_{i}=\left\|\boldsymbol{y}-\mathbf{X}_{i} \hat{\boldsymbol{\rho}}_{i}\right\|_{2} /\left\|\hat{\boldsymbol{\rho}}_{i}\right\|_{2}$ is used to classify the test image $\boldsymbol{y}$ by utilizing the discrimination information contained in $\left\|\hat{\boldsymbol{\rho}}_{i}\right\|_{2}$, where $\hat{\boldsymbol{\rho}}_{i}$ is the coefficient vector associated with class $i$. Finally, the test sample is designated to the class that has the least regularized residual.

\subsection{Low-rank representation (LRR)}
\label{sec-2.2}
Low-rank matrix recovery technique is used in our proposed method to recover a clean data matrix, so we investigate its formulation for the purpose of completeness. The corrupted training samples $\mathbf{X}$ can be decomposed into $\mathbf{D}+\mathbf{E}$ by RPCA \cite{wright2009robust}, in which $\mathbf{D}$ is the clean low rank matrix and $\mathbf{E}$ is the associated sparse error matrix. The rank of matrix $\mathbf{D}$ is minimized by RPCA, meanwhile, $\|\mathbf{E}\|_{0}$ is reduced for the sake of deriving the low-rank approximation of $\mathbf{X}$. The original formulation of RPCA is formulated as, 
\begin{equation}\label{equ-4}
\min _{\mathbf{D}, \mathbf{E}} \operatorname{rank}(\mathbf{D})+\lambda\|\mathbf{E}\|_{0} \text { s.t. } \mathbf{X}=\mathbf{D}+\mathbf{E}
\end{equation}
where $\|\cdot\|_{0}$ denotes the $\ell_0$-norm. Eq. (\ref{equ-4}) is NP-hard as well as highly nonconvex, it is relaxed by replacing the $\ell_0$-norm with the $\ell_1$-norm and the rank function with the nuclear norm. The new optimization problem is more tractable as follows:
\begin{equation}\label{equ-5}
\min _{\mathbf{D}, \mathbf{E}}\|\mathbf{D}\|_{*}+\lambda\|\mathbf{E}\|_{1} \text { s.t. } \mathbf{X}=\mathbf{D}+\mathbf{E}
\end{equation}

The limitation of RPCA is that it assumes that all the column vectors in $\mathbf{X}$ are drawn from a single low-rank subspace \cite{liu2012robust}. This hypothesis is not general and reasonable because face images usually reside on a union of multiple subspaces. A modified rank optimization problem in LRR is presented by Liu \etal \cite{liu2010robust,liu2012robust} defined as follows:         
\begin{equation}\label{equ-6}
\min _{\mathbf{Z}, \mathbf{E}}\|\mathbf{Z}\|_{*}+\lambda\|\mathbf{E}\|_{l} \text { s.t. } \mathbf{X}=\mathbf{X} \mathbf{Z}+\mathbf{E}
\end{equation}
where the representation matrix of $\mathbf{X}$ is denoted by $\mathbf{Z}$ and $\|\cdot\|_{l}$ is a certain regularization strategy for expressing different corruptions. The inexact augmented Lagrange multipliers (ALM) algorithm \cite{lin2010augmented} is employed to efficiently solve the above optimization problem (\ref{equ-6}). After the optimal solution $\mathbf{Z}^{*}$ is obtained, the recovered clean data matrix can be acquired from the corrupted data matrix $\mathbf{X}$ by $\mathbf{XZ}^{*}$.

\section{Proposed method}
\label{sec-3}
In order to deal with the problem of robust FR, where training and test images might be simultaneously corrupted by outlier and cannot be well solved by CRC, a discriminative low-rank representation method is proposed in this section. Our method can recover a clean and discriminative dictionary from the highly corrupted training images. To handle the corruption appeared in the test samples, a low-rank projection matrix is learned to project test samples onto their corresponding underlying subspaces and obtain the new clean test samples. In the testing stage, CRC is exploited to classify the corrected test samples.

\subsection{Discriminative low-rank representation for matrix recovery}
\label{sec-3.1}
The low-rank matrix recovery techniques can be utilized to improve the recognition accuracy of CRC because the problems brought by corrupted training samples can be alleviated. The recovered clean dictionary with a better representation ability can be obtained as $\mathbf{XZ}^{*}$ from the original matrix $\mathbf{X}$ by solving (\ref{equ-6}). One fact is that face images from different people share similarities due to the location of eyes, mouth, etc., so the drawback is that discriminative information is not contained by $\mathbf{XZ}^{*}$ and it is not suitable for classification. Inspired by \cite{ramirez2010classification}, we propose a DLRR scheme for matrix recovery, different from LRR, the incoherence between different subjects in $\mathbf{XZ}^{*}$ is promoted. Consequently, the introduction of such incoherence would make the derived low-rank matrix from different subjects as independent as possible. The discrimination capacity is improved while the commonly shared features are suppressed.

A set of training images with corruptions are set as the original data matrix $\mathbf{X}=\left[\mathbf{X}_{1}, \mathbf{X}_{2}, \ldots, \mathbf{X}_{N}\right]$, where $\mathbf{X}_i$ is composed of the training images from class $i$. As mentioned above, the data matrix $\mathbf{X}$ can be decomposed into a low-rank matrix $\mathbf{D}=\left[\mathbf{D}_{1}, \mathbf{D}_{2}, \ldots, \mathbf{D}_{N}\right]$ and the sparse error matrix $\mathbf{E}$ by the formulation in (\ref{equ-6}), where $\mathbf{D}_{i}=\mathbf{X}_{i} \mathbf{Z}_{i}$ represents the clean data matrix from subject $i$. A regularization term $\left\|\mathbf{D}_{j}^{T} \mathbf{D}_{i}\right\|_{F}^{2}$ which sums the Frobenius norms of each pair of low-rank matrix $\mathbf{D}$ is added to the original LRR formulation to improve the independence of different classes. If the value of $\left\|\mathbf{D}_{j}^{T} \mathbf{D}_{i}\right\|_{F}^{2}$ is as small as possible, then the between-class independence can be achieved in our method. The new optimization problem is formulated as follows,
\begin{equation}\label{equ-7}
\begin{array}{cl}\underset{\mathbf{Z}_i,\mathbf{E}_i}{\textrm{min}} &\ {\left\|\mathbf{Z}_{i}\right\|_{*}+\lambda\left\|\mathbf{E}_{i}\right\|_{2,1}+\frac{\eta}{2} \sum_{j=1, i \neq j}^{N}\left\|\left(\mathbf{X}_{i} \mathbf{Z}_{i}\right)^{T} \mathbf{X}_{j} \mathbf{Z}_{j}\right\|_{F}^{2}} \\ {\text { s.t. }} & {\mathbf{X}_{i}=\mathbf{X}_{i} \mathbf{Z}_{i}+\mathbf{E}_{i}}\end{array}
\end{equation}
where $\lambda$ is scalar parameter. The last term promotes the structural incoherence of different subjects, which is penalized by the scalar parameter $\eta$ balancing the low-rank matrix decomposition and discriminative features. When draw a comparison between (\ref{equ-7}) and (\ref{equ-6}), the regularization term is utilized to provide improved discriminative ability, which can enforce more training samples from the correct subject to represent the test samples. We use the $\ell_{2,1}$-norm to encourage the columns of error term matrix $\mathbf{E}$ which represents extreme noise to be zero, the other extra regularization is unnecessary to be used on $\mathbf{E}$ since $\mathbf{E}$ is sparse. As a result, our new formulation (\ref{equ-7}) can fully explore the discrimination capacity contained in the original corrupted training images.

\begin{figure}[h]
\centering
\includegraphics[trim={0mm 0mm 0mm 0mm},clip, width = .8\textwidth]{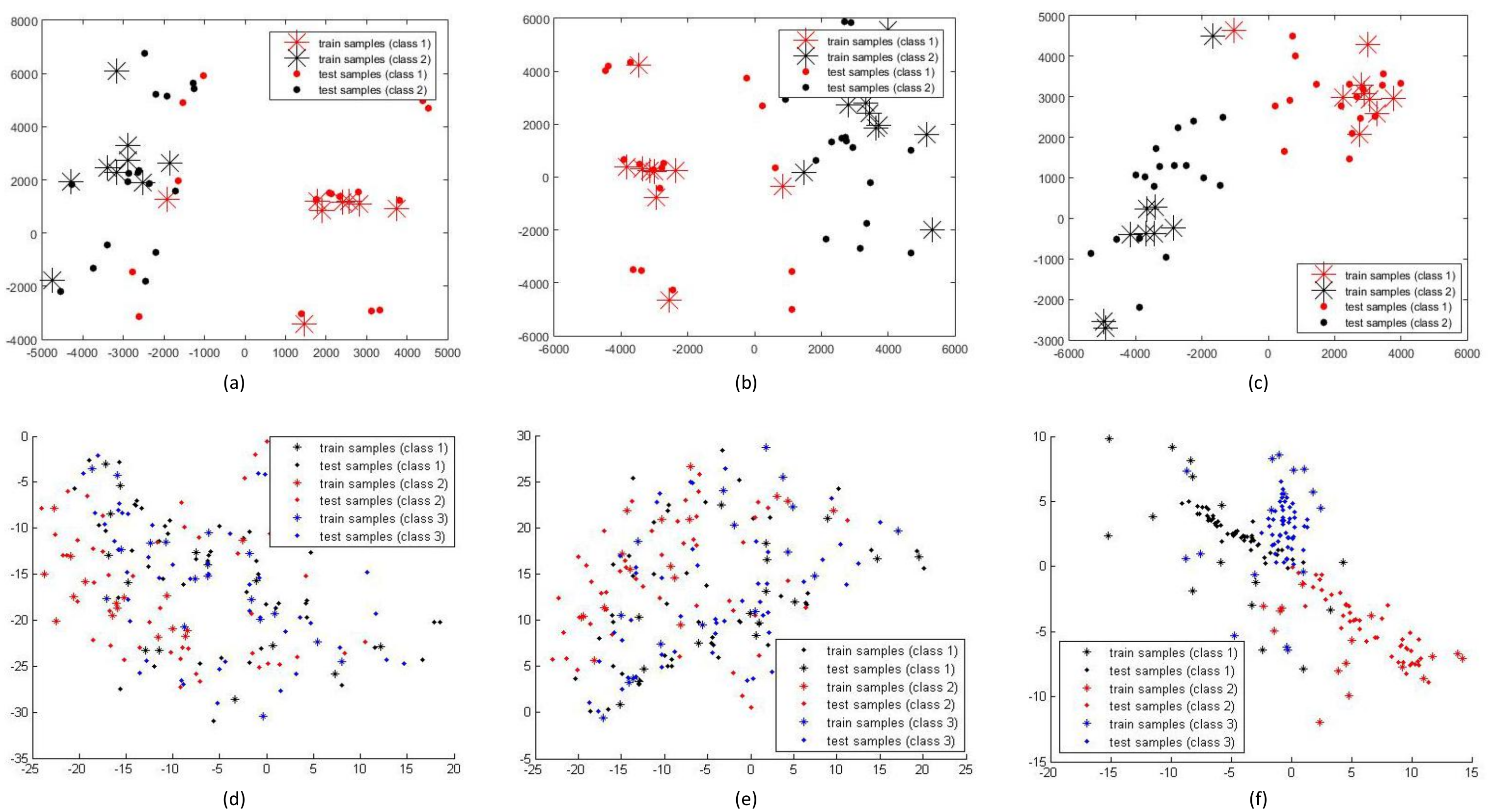}
\caption{Data distributions for different classes represented by different colors. The training and test data are projected onto the first two eigenvectors of the covariance matrices of (a) the original data matrix $\mathbf{X}$, (b) the LR matrix $\mathbf{XZ}^*$ without structural incoherence, (c) the LR matrix $\mathbf{XZ}^*$ with structural incoherence. When the training and test data are spanned by three different classes, the corresponding plots for (a), (b) and (c) are shown in (d), (e) and (f), respectively.}
\label{fig:data_distri}
\end{figure}

The discriminatory ability of the DLRR scheme in classifying face samples from different subjects is illustrated in Fig.~\ref{fig:data_distri}. Similar to former Eigenfaces and CRC-based methods, we use the training and test face images from two different subjects in the AR face database and then project them onto the first two eigenvectors of the covariance matrix of the original data matrix $\mathbf{X}$ as shown in Fig.~\ref{fig:data_distri} (a), then the same data is projected onto the subspace which is derived by $\mathbf{XZ}^{*}$ without structural incoherence, the result is shown in Fig.~\ref{fig:data_distri} (b), and Fig.~\ref{fig:data_distri} (c) is the result of our proposed method. It is obvious that the distinction between the data projected onto $\mathbf{XZ}^{*}$ with and without structural incoherence are both improved compared with Fig.~\ref{fig:data_distri} (a). However, compared Fig.~\ref{fig:data_distri} (b) with Fig.~\ref{fig:data_distri} (c), we can find that the within-class scatter in Fig.~\ref{fig:data_distri} (c) is smaller than that of in Fig.~\ref{fig:data_distri} (b), and thus a better discriminative ability can be obtained by using our proposed approach. We also choose other images from three different classes in the Extended Yale B face database to do the same experiment as described above, as shown in the second row of Fig.~\ref{fig:data_distri}. We plot their corresponding 2D subspaces spanned by the first two eigenvectors in Figs.~\ref{fig:data_distri} (d), (e) and (f), respectively. It is worth noting that our method, \ie Fig.~\ref{fig:data_distri} (f), exhibits desirable discrimination capacity, while the distinction between the data projected onto the original data matrix $\mathbf{X}$ and the low-rank matrix $\mathbf{XZ}^{*}$ are observed to be degraded. Hence, better representation ability can be achieve by utilizing our derived LR matrix.

\subsection{Optimization via ALM}
\label{sec-3.2}
Our proposed optimization problem (\ref{equ-7}) is solved by ALM \cite{lin2010augmented} in this section. Firstly, we introduce an auxiliary variable $\mathbf{J}_{i}$ to make the optimization problem (\ref{equ-7}) solvable, the new equivalent optimization problem is given as follows,
\begin{equation}\label{equ-8}
\begin{array}{ll}{\min _{\mathbf{Z}_{i}, \mathbf{E}_{i} \mathbf{J}_{i}}\left\|\mathbf{Z}_{i}\right\|_{*}+\lambda\left\|\mathbf{E}_{i}\right\|_{2,1}+\frac{\eta}{2} \sum_{j=1 ; i \neq j}^{N}\left\|\left(\mathbf{X}_{j} \mathbf{Z}_{j}\right)^{T} \mathbf{X}_{i} \mathbf{J}_{i}\right\|_{F}^{2}} \\ {\text { s.t. } \ \mathbf{X}_{i}=\mathbf{X}_{i} \mathbf{Z}_{i}+\mathbf{E}_{i}, \mathbf{Z}_{i}=\mathbf{J}_{i}}\end{array}
\end{equation}

The transformed augmented Lagrangian function is formulated as an unconstrained optimization problem,
\begin{equation}\label{equ-9}
\begin{array}{l}{L\left(\mathbf{Z}_{i}, \mathbf{E}_{i}, \mathbf{J}_{i}, \mathbf{Y}_{1}, \mathbf{Y}_{2},\mu\right)} \\ {=\left\|\mathbf{Z}_{i}\right\|_{*}+\lambda\left\|\mathbf{E}_{i}\right\|_{2,1}+\frac{\eta}{2} \sum_{j=1, i \neq j}^{N} \| \mathbf{Z}_{j}^{T} \mathbf{X}_{j}^{T} \mathbf{X}_{i} \mathbf{J}_{i} \|_{F}^{2}+\left\langle \mathbf{X}_{i}-\mathbf{X}_{i} \mathbf{Z}_{i}-\mathbf{E}_{i}, \mathbf{Y}_{1}\right\rangle} \\
{+\left\langle \mathbf{Z}_{i}-\mathbf{J}_{i}, \mathbf{Y}_{2}\right\rangle +\frac{\mu}{2}\left(\left\|\mathbf{X}_{i}-\mathbf{X}_{i} \mathbf{Z}_{i}-\mathbf{E}_{i}\right\|_{F}^{2}+\left\|\mathbf{Z}_{i}-\mathbf{J}_{i}\right\|_{F}^{2}\right)}\end{array}
\end{equation}
where $\mathbf{Y}_{1}$ and $\mathbf{Y}_{2}$ are Lagrangian multipliers, and $\mu>0$ is used as a penalty parameter. The optimization problem (\ref{equ-9}) can be rewritten by some simple algebra as follows,
\begin{equation}\label{equ-10}
\begin{array}{l}{L\left(\mathbf{Z}_{i}, \mathbf{E}_{i}, \mathbf{J}_{i}, \mathbf{Y}_{1}, \mathbf{Y}_{2},\mu\right)} \\ {=\left\|\mathbf{Z}_{i}\right\|_{*}+\lambda\left\|\mathbf{E}_{i}\right\|_{2,1}+\frac{\eta}{2} \sum_{j=1, i \neq j}^{N}\| \mathbf{Z}_{j}^{T} \mathbf{X}_{j}^{T} \mathbf{X}_{i} \mathbf{J}_{i} \|_{F}^{2}+f\left(\mathbf{Z}_{i}, \mathbf{J}_{i}, \mathbf{E}_{i}, \mathbf{Y}_{1}, \mathbf{Y}_{2}, \mu\right)}\end{array}
\end{equation}
where $f\left(\mathbf{Z}_{i}, \mathbf{J}_{i}, \mathbf{E}_{i}, \mathbf{Y}_{1}, \mathbf{Y}_{2}, \mu\right)=\frac{\mu}{2}\left(\left\|\mathbf{X}_{i}-\mathbf{X}_{i} \mathbf{Z}_{i}-\mathbf{E}_{i}+\frac{\mathbf{Y}_{1}}{\mu}\right\|_{F}^{2}+\left\|\mathbf{Z}_{i}-\mathbf{J}_{i}+\frac{\mathbf{Y}_{2}}{\mu}\right\|_{F}^{2}\right)$

The variables $\mathbf{Z}_i$, $\mathbf{J}_i$, $\mathbf{E}_i$ could be iteratively updated. Two variables are fixed at each iteration to update the remaining one. The detailed updating schemes are showed as follows in each step.

\begin{enumerate}[1)]
	\item Updating $\mathbf{Z}_i$
	
	Updating $\mathbf{Z}_i$ by minimizing $L\left(\mathbf{Z}_{i}, \mathbf{E}_{i}^{k}, \mathbf{J}_{i}^{k}, \mathbf{Y}_{1}^{k}, \mathbf{Y}_{2}^{k},\mu_k\right)$ is equivalent to minimizing the following unconstrained minimization function for $\sigma=\left\|\mathbf{A}_{i}\right\|_{2}^{2}$.
	
	$\arg \min \left\|\mathbf{Z}_{i}\right\|_{*}+\left\langle \mathbf{Z}_{i}-\mathbf{Z}_{i}^{k}, \nabla_{\mathbf{Z}_{i}} f\left(\mathbf{Z}_{i}^{k}, \mathbf{J}_{i}^{k}, \mathbf{E}_{i}^{k}, \mathbf{Y}_{1}^{k}, \mathbf{Y}_{2}^{k}, \mu_{k}\right)\right\rangle$
	$+\frac{\mu \sigma}{2}\left\|\mathbf{Z}_{i}-\mathbf{Z}_{i}^{k}\right\|_{F}^{2}$
	where $\nabla_{\mathbf{Z}_{i}} f\left(\mathbf{Z}_{i}^{k}\right)=\mu\left[-\mathbf{X}_{i}^{T}\left(\mathbf{X}_{i}-\mathbf{X}_{i} \mathbf{Z}_{i}^{k}-\mathbf{E}_{i}^{k}+\frac{\mathbf{Y}_{1}^{k}}{\mu_{k}}\right)+\left(\mathbf{Z}_{i}^{k}-\mathbf{J}_{i}^{k}+\frac{\mathbf{Y}_{2}^{k}}{\mu_{k}}\right)\right]$
	
The above problem has the following closed-form solution,
	\begin{equation}\label{equ-11}
	\mathbf{Z}_{i}^{k+1}=\arg \min \left\|\mathbf{Z}_{i}\right\|_{*}+\frac{\mu \sigma}{2}\left\|\mathbf{Z}_{i}-\left(Z_{i}^{k}-\frac{\nabla_{\mathbf{Z}} f\left(\mathbf{Z}_{i}^{k}\right)}{\mu \sigma}\right)\right\|_{F}^{2}
	\end{equation}
	
	\item Updating $\mathbf{J}_i$
	
	Now by fixing $\mathbf{Z}_i$, $\mathbf{E}_i$, $\mathbf{Y}_1$ and $\mathbf{Y}_2$, we optimize the variable $\mathbf{J}_i$ for class $i$. The updating scheme of $\mathbf{J}_i$ is as follows:
	
	$\mathbf{J}_{i}^{k+1}=\arg \min _{\mathbf{J}_{i}} L\left(\mathbf{Z}_{i}^{k+1}, \mathbf{E}_{i}^{k}, \mathbf{J}_{i}, \mathbf{Y}_{1}^{k}, \mathbf{Y}_{2}^{k},\mu_k\right)$
	
	Then the solution to the above problem can be solved by solving the partial derivatives of $L$  w.r.t. $\mathbf{J}_i$ and setting it to be zero, and the solution is given by, 
	\begin{equation}\label{equ-12}
\mathbf{J}_{i}^{k+1}=\left(\eta \sum_{j=1, i \neq j}^{N} \mu \mathbf{B}_{j}^{T} \mathbf{B}_{j}+\mu_{k} \mathbf{I}\right)^{-1}\left(\mu_{k} \mathbf{Z}_{i}^{k+1}+\mathbf{Y}_{2}^{k}\right)
	\end{equation}
	where $\mathbf{B}_{j}=\mathbf{Z}_{j}^{T} \mathbf{X}_{j}^{T} \mathbf{X}_{i}$
	
	\item Updating $\mathbf{E}_i$
	
	By fixing $\mathbf{Z}_i$, $\mathbf{J}_i$, $\mathbf{Y}_1$ and $\mathbf{Y}_2$, we update the error matrix $\mathbf{E}_i$ for subject $i$ as follows,
	\begin{equation}\label{equ-13}
	\begin{array}{l}{\mathbf{E}_{i}^{k+1}=\min _{\mathbf{E}_{i}} \lambda\left\|\mathbf{E}_{i}\right\|_{2,1}+\left\langle \mathbf{X}_{i}-\mathbf{X}_{i} \mathbf{Z}_{i}^{k+1}-\mathbf{E}_{i}, \mathbf{Y}_{1}^{k}\right\rangle+\frac{\mu_{k}}{2}\left\|\mathbf{X}_{i}-\mathbf{X}_{i} \mathbf{Z}_{i}^{k+1}-\mathbf{E}_{i}\right\|_{F}^{2}} \\ {=\min _{\mathbf{E}_{i}} \lambda\left\|\mathbf{E}_{i}\right\|_{2,1}+\frac{\mu_{k}}{2} \operatorname{tr}\left[\left(\mathbf{X}_{i}-\mathbf{X}_{i} \mathbf{Z}_{i}^{k+1}-\mathbf{E}_{i}\right)^{T}\left(\mathbf{X}_{i}-\mathbf{X}_{i} \mathbf{Z}_{i}^{k+1}-\mathbf{E}_{i}\right)\right.} \\ {\left.+\frac{2 \mathbf{Y}_{1}^{k}}{\mu_{k}}\left(\mathbf{X}_{i}-\mathbf{X}_{i} \mathbf{Z}_{i}^{k+1}-\mathbf{E}_{i}\right)^{T}\right]} \\ {=\min _{\mathbf{E}_{i}} \frac{\lambda}{\mu_{k}}\left\|\mathbf{E}_{i}\right\|_{2,1}+\frac{1}{2}\left\|\mathbf{E}_{i}-\left(\mathbf{X}_{i}-\mathbf{X}_{i} \mathbf{Z}_{i}^{k+1}+\frac{\mathbf{Y}_{1}^{k}}{\mu_{k}}\right)\right\|_{F}^{2}}\end{array}
	\end{equation}
\end{enumerate}

Algorithm 1 summaries the whole detailed procedures for solving the optimization problem (\ref{equ-10}).

\begin{algorithm}[t]
	\begin{algorithmic}
		\vspace{0.03in}
		\STATE\textbf{Input:} Training data matrix $\mathbf{X}$ for $N$ classes, and parameters $\lambda>0$, $\eta>0$
		\STATE\ \ \ \ 1. for $i=1 ; i<N ; i++$
		\STATE\ \ \ \ 2. Initialize: $\mathbf{Z}_{i}=\mathbf{J}_{i}=\mathbf{E}_{i}=\mathbf{Y}_{1}=\mathbf{Y}_{2}=0, \mathbf{Z}_{i}=\mathbf{I}, \mu=10^{-6}, \rho=1.1, \mu_{\max }=10^{10}, \varepsilon=10^{-3}, k=0$
		\STATE\ \ \ \ 3. while not converged, $k \leq$maxiter do
		\STATE\ \ \ \ 4. update $\mathbf{Z}_i$ according to (\ref{equ-11});
		\STATE\ \ \ \ 5. update $\mathbf{J}_i$ according to (\ref{equ-12});
		\STATE\ \ \ \ 6. update $\mathbf{E}_i$ according to (\ref{equ-13});
		\STATE\ \ \ \ 7. update the Lagrange multipliers
		\STATE\ \ \ \ \ \ \ $\mathbf{Y}_{1}^{k+1}=\mathbf{Y}_{1}^{k}+\mu_{k}\left(\mathbf{X}_{i}-\mathbf{X}_{i} \mathbf{Z}_{i}^{k+1}-\mathbf{E}_{i}^{k+1}\right)$;
		\STATE\ \ \ \ \ \ \ $\mathbf{Y}_{2}^{k+1}=\mathbf{Y}_{2}^{k}+\mu_{k}\left(\mathbf{Z}_{i}^{k+1}-\mathbf{J}_{i}^{k+1}\right)$;
		\STATE\ \ \ \ 8. update the parameter $\mu$ by $\mu_{k+1}=\min \left(\rho \mu_{k}, \mu_{\max }\right)$;
		\STATE\ \ \ \ 9. check the convergence conditions
		\STATE\ \ \ \ \ \ \ $\left\|\mathbf{X}_{i}-\mathbf{X}_{i} \mathbf{Z}_{i}^{k+1}-\mathbf{E}_{i}^{k+1}\right\|_{\infty}<\varepsilon$ and $\left\|\mathbf{Z}_{i}^{k+1}-\mathbf{J}_{i}^{k+1}\right\|_{\infty}<\varepsilon$
		\STATE\ \ \ \ 10. update $k: k \leftarrow k+1$
		\STATE\ \ \ \ 11. end while
		\STATE\ \ \ \ 12. end for
		\STATE\ \ \ \ 13. an optimal solution $\mathbf{Z}^*$
		\STATE\textbf{Output:} the recovered clean data matrix $\mathbf{D}=\mathbf{XZ}^*$.
		\vspace{0.03in}
	\end{algorithmic}
	\caption{Solving Problem (\ref{equ-8}) by ALM}
	\label{alg1}
\end{algorithm}

\begin{figure}[h]
\centering
\includegraphics[trim={0mm 0mm 0mm 0mm},clip, width = .8\textwidth]{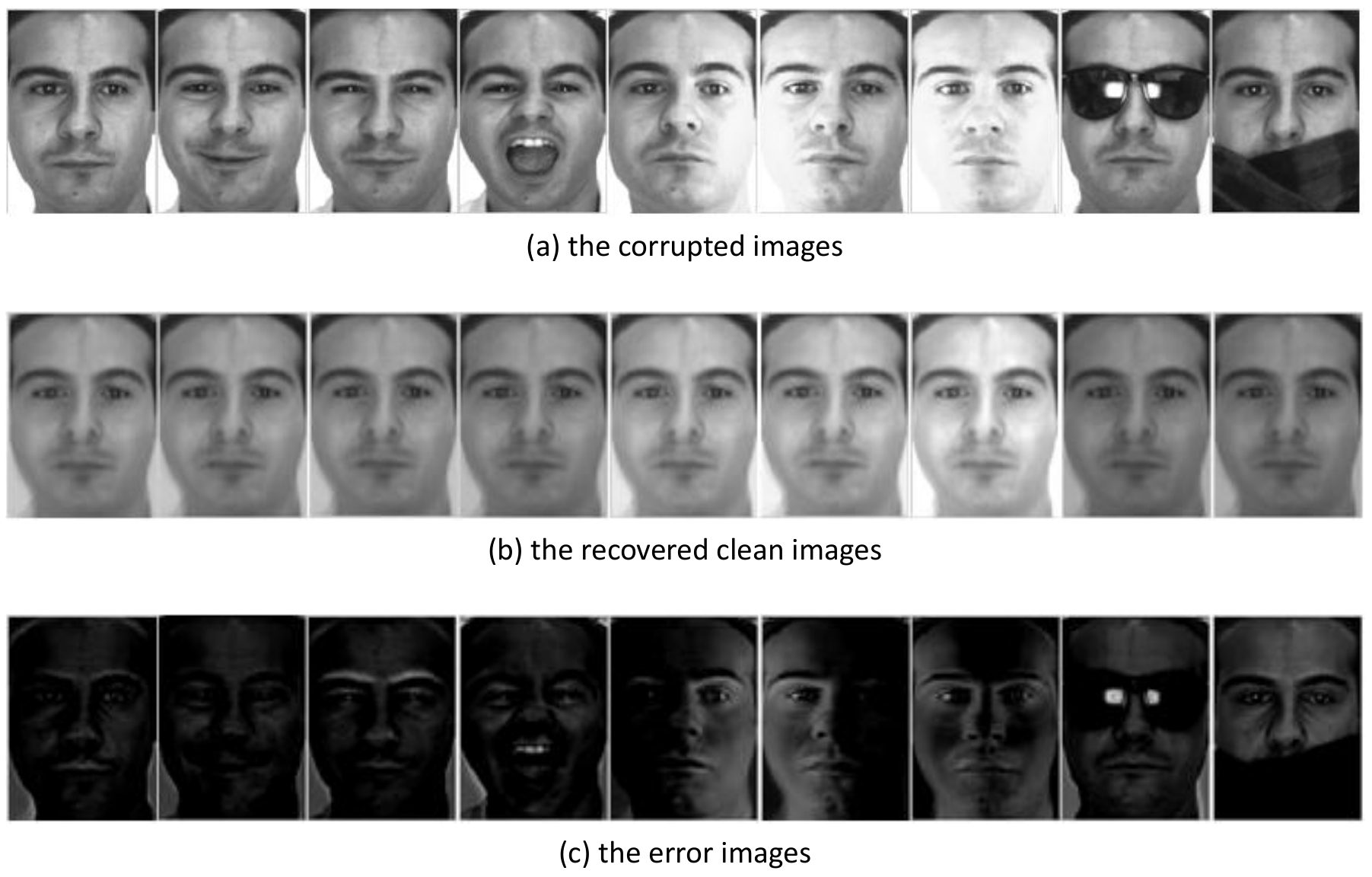}
\caption{Recovery results of DLRR.}
\label{fig:tr_recover}
\end{figure}

An example is given in Fig.~\ref{fig:tr_recover} to intuitively display the recovery results of DLRR for matrix recovery, the corrupted images are successfully separated into the recovered clean images and the error images. Some training samples from one subject in AR database with illumination variations, pose changes, and occlusions are shown in Fig.~\ref{fig:tr_recover} (a). The recovered clean samples and the corresponding sparse errors are shown in Figs.~\ref{fig:tr_recover} (b) and (c), respectively.

\subsection{Low-rank projection matrix}
\label{sec-3.3}
In the testing phase, to handle the possible occlusion variations appeared in the test samples, which could degrade the performance of CRC. Motivated by Bao \etal \cite{bao2012inductive}, we try to find a low-rank projection matrix which can project the new corrupted samples onto their corresponding underlying subspaces.
  
After the original corrupted samples $\mathbf{X}=\left[\mathbf{X}_{1}, \ldots, \mathbf{X}_{N}\right] \in \mathbb{R}^{m \times n}$ are successfully separated into the recovered matrix $\mathbf{Y}=\left[\mathbf{X}_{1}\mathbf{Z}_{1}, \ldots, \mathbf{X}_{N}\mathbf{Z}_{N}\right] \in \mathbb{R}^{m \times n}$ and the sparse error matrix. The matrix $\mathbf{Y}$ can be seen as the set of principal components obtained from the matrix $\mathbf{X}$, so a linear projection $\mathbf{P}$ can be studied between $\mathbf{X}$ and $\mathbf{Y}$. Then any data sample $\boldsymbol{x}$ can be projected onto its underlying subspace to get the recovery results as $\mathbf{P}\boldsymbol{x}$. Based on the assumption that the recovery result is considered to be drawn from a union of multiple low-rank subspaces, we could hypothesize that $\mathbf{P}$ is a low-rank matrix and the optimization problem can be formulated as follows,
\begin{equation}\label{equ-14}
\min _{\mathbf{P}} \operatorname{rank}(\mathbf{P}) \quad \text { s.t. } \quad \mathbf{Y}=\mathbf{P X}
\end{equation}

As mentioned in Section \ref{sec-2}, a convex relaxation of the optimization problem \ref{equ-14} is proposed by replacing the rank function with the nuclear norm which can decrease the computational complexity. The convex optimization problem is formulated as,
\begin{equation}\label{equ-15}
\min _{\mathbf{P}}\|\mathbf{P}\|_{*} \quad \text { s.t. } \quad \mathbf{Y}=\mathbf{P X}
\end{equation}

$\mathbf{P}^{*}=\mathbf{Y X}^{+}$ is formulated as the uniqueness of the minimizer for the above problem (\ref{equ-15}) under the hypothesis that $\mathbf{P} \neq 0$ and $\mathbf{Y}=\mathbf{PX}$ has feasible solution, where $\mathbf{X}^{+}$ is the pseudo-inverse of $\mathbf{X}$. The new test sample $\boldsymbol{y}$ can be recovered by $\mathbf{P}^{*}\boldsymbol{y}$ after obtained the optimal solution $\mathbf{P}^{*}$.

We outline the detailed procedures of collaborative representation-based classification by discriminative low-rank representation in Algorithm~\ref{alg2}.

\begin{algorithm}[t]
	\begin{algorithmic}
		\vspace{0.03in}
		\STATE\textbf{Input:} Training data matrix $\mathbf{X}=\left[\boldsymbol{x}_{1}, \boldsymbol{x}_{2}, \ldots, \boldsymbol{x}_{n}\right] \in \mathbb{R}^{m \times n}$ for $N$ classes, a test sample $\boldsymbol{y} \in \mathrm{R}^{m}$ and parameters $\lambda>0, \eta>0, \beta>0$
		\STATE\ \ \ \ 1. for $i=1 ; i<N ; i++$
		\STATE\ \ \ \ 2. Find an optimal solution ($\mathbf{Z}^*$) of the following optimization problem by Algorithm \ref{alg1}:
		\STATE\ \ \ \ \ \ $\min _{\mathbf{Z}_{i}, \mathbf{E}_{i}, \mathbf{J}_{i}}\left\|\mathbf{Z}_{i}\right\|_{*}+\lambda\left\|\mathbf{E}_{i}\right\|_{2,1}+\frac{\eta}{2} \sum_{j=1, i \neq j}^{N}\left\|\left(\mathbf{X}_{j} \mathbf{Z}_{j}\right)^{T} \mathbf{X}_{i} \mathbf{Z}_{i}\right\|_{F}^{2}$
		\STATE\ \ \ \ \ \ s.t.$\quad \mathbf{X}_{i}=\mathbf{X}_{i} \mathbf{Z}_{i}+\mathbf{E}_{i}$
		\STATE\ \ \ \ 3. end for
		\STATE\ \ \ \ 4. Correct a test sample $\boldsymbol{y}$:
		\STATE\ \ \ \ \ \ $\mathbf{Y}=\left[\mathbf{X}_{1} \mathbf{Z}_{1}, \ldots, \mathbf{X}_{N} \mathbf{Z}_{N}\right]$
		\STATE\ \ \ \ \ \ $\mathbf{P}^{*}=\mathbf{Y} \mathbf{X}^{+} ;$
		\STATE\ \ \ \ \ \ $\boldsymbol{y}_{p}=\mathbf{P}^{*} \boldsymbol{y}$
		\STATE\ \ \ \ 5. Classify $\boldsymbol{y}_p$ by CRC
		\STATE\ \ \ \ \ \ $\min _{\boldsymbol{\rho}}\left\{\left\|\boldsymbol{y}_{p}-\mathbf{X} \cdot \boldsymbol{\rho}\right\|_{2}^{2}+\beta\|\boldsymbol{\rho}\|_{2}^{2}\right\}$
		\STATE\ \ \ \ 6. for $i=1 ; i<N ; i++$
		\STATE\ \ \ \ 7. $\boldsymbol{e}_{i}=\left\|\boldsymbol{y}_{p}-\mathbf{X}_{i} \boldsymbol{\rho}_{i}\right\|_{2} /\left\|\boldsymbol{\rho}_{i}\right\|_{2}$
		\STATE\ \ \ \ 8. end for
		\STATE\textbf{Output:} identity $(\boldsymbol{y})=\arg \min _{i}\left\{\boldsymbol{e}_{i}\right\}$
		\vspace{0.03in}
	\end{algorithmic}
	\caption{Collaborative representation-based classification by discriminative low-rank representation}
	\label{alg2}
\end{algorithm}

\section{Experiments}
\label{sec-4}
In this section, the performance of our proposed DLRR-CR is evaluated on two face databases: AR \cite{martinez1998ar} and Extended Yale B \cite{georghiades2001few} databases. The most face images chosen for face recognition are with variations in illumination, expression and corruption etc., and compare the performance of our method with the state-of-the-art methods, including SRC \cite{wright2008robust}, CRC \cite{zhang2011sparse}, LRC (linear regression classification) \cite{naseem2010linear} and NN. To demonstrate the discriminating ability of the additional item $\left\|\mathbf{D}_{j}^{T} \mathbf{D}_{i}\right\|_{F}^{2}$ in (\ref{equ-7}), the LRR-CRC is implemented in our experiments. The DLRR-based SRC algorithm is also implemented denoted by DLRR-SRC to compare the effectiveness of SRC and CRC in the testing phase. In our experiments, high dimensionality of face images will lead to high computation complexity, so PCA is used as the dimensionality reduction method before testing. In our methods, the new learned eigenspace is spanned by the eigenvectors of the covariance  matrix of the LR matrix $\mathbf{D}$ with structural incoherence. We choose the feature dimensions of 25, 50, 75, 100, 200 and 300. The method ALM \cite{yang2011alternating} is used to solve the $\ell_1$-norm problem, and the regularization parameter in ALM is set to 0.001.

\begin{figure}[h]
\centering
\includegraphics[trim={0mm 0mm 0mm 0mm},clip, width = .8\textwidth]{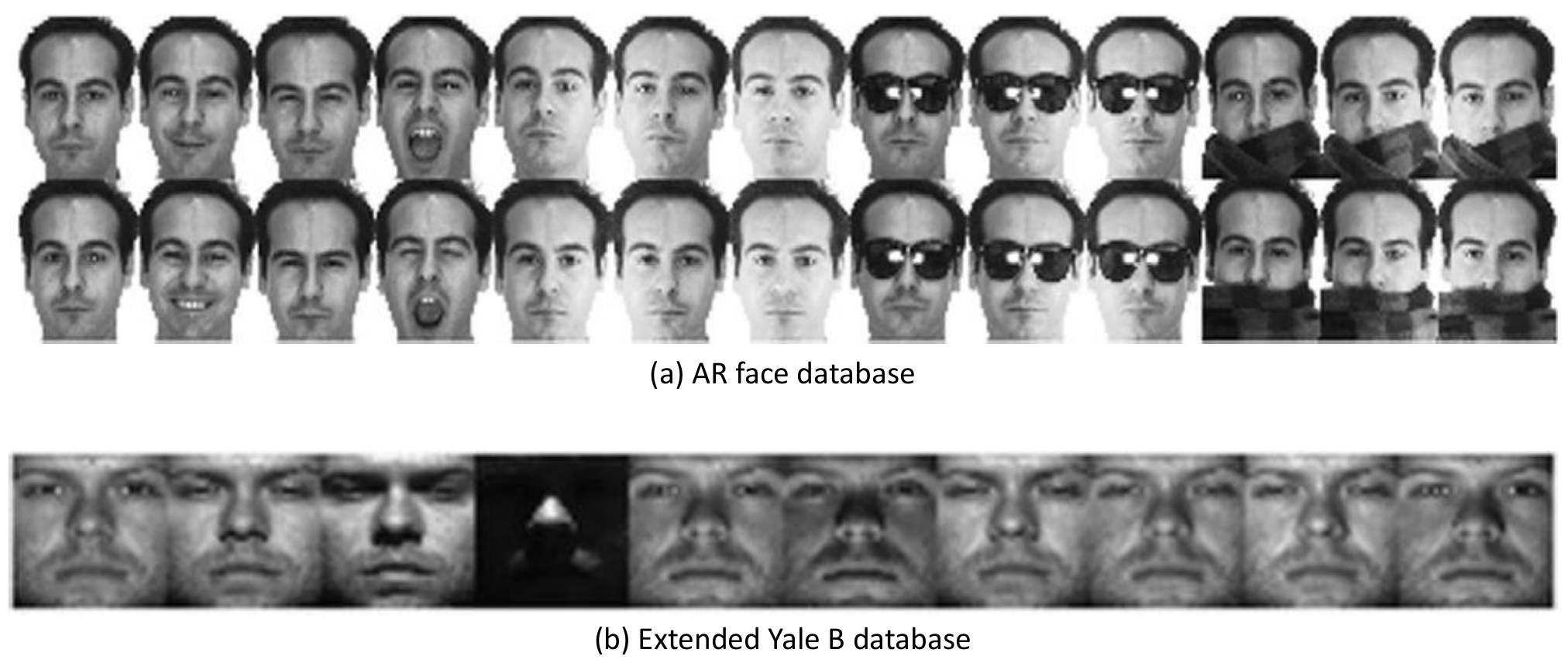}
\caption{Example training and test images used in our experiments.}
\label{fig:exam_img}
\end{figure}

\subsection{AR database}
\label{sec-4.1}
The AR database \cite{martinez1998ar} contains over 4,000 frontal images from 126 subjects. In our experiments a subset from the AR database that contains 50 male subjects and 50 female subjects are used. There are 26 face images available for each subject, divided into two sessions, under the variations of expression, illumination, and disguise. In each session, there are 7 clean images with illumination and expressions variations, 3 images in sunglasses and the remaining 3 images in scarves disguise. All face images are cropped to 165$\times$120 pixels and then converted into grayscale before training and testing. Some images from the first subject in the AR database are shown in Fig.~\ref{fig:exam_img} (a). We consider the following three scenarios to validate the performance of DLRR-CR as in \cite{chen2012low}.
\begin{enumerate}[1)]
	\item Sunglasses: We consider the situation in which training and test images corrupted by sunglasses simultaneously. The presence of sunglasses produce about 20\% occlusion of the frontal face image. From session one, we use eight training images, 7 neutral images plus one image with sunglasses. We use twelve test images, all non-occluded images from session two plus the rest of the face images with sunglasses.
	
	\item Scarf: We consider the situation in which training and test images corrupted by disguise simultaneously due to scarf, which occlude about 40\% of the frontal face image. A similar choice of training and testing set is applied as above. From the first session, we use 7 neutral plus one with scarf for training. 7 non-occluded from the other session plus the remaining images with scarves for testing.
	
	\item Sunglasses+Scarf: In the final situation, we consider the most challenging case which training and test images occluded by mixed corruption due to sunglasses and scarves. From the first session, we use 7 neutral plus one with sunglasses and the other with a scarf for training. 7 non-occluded from the other session add the remaining images for testing.
\end{enumerate}

For fair comparison, the feature dimensions are reduced to the same size for all methods. The regularization parameters in our method are set as $\beta$=1.1, $\eta=0.001$ and $\lambda$=0.02. The recognition accuracy of the three situations are given in Tables~\ref{tab-1}-\ref{tab-3}. From the results we can see that our method almost exceeds the other methods in each situation, which demonstrates the superiority of our method over the other approaches in dealing with real disguise.

\begin{table}[]
	\caption{Recognition accuracy (\%) of different methods on the AR database with the occlusion of sunglasses.}
	\label{tab-1}
	\centering
	\begin{tabular}{ccccccc}
		\hline
		Dim & 25& 50& 75& 100& 200& 300\\
		\hline
		DLRR-CR & \textbf{65.58} & \textbf{82.00} & 87.33          & \textbf{90.00} & \textbf{92.00} & 91.75          \\
		LRR-CR  & 61.25          & 81.58          & \textbf{87.50} & 89.50          & 90.67          & 90.58          \\
		LRR-CRC & 54.75          & 75.67          & 83.08          & 86.67          & 91.33          & \textbf{92.08} \\
		CRC     & 52.08          & 73.67          & 80.25          & 84.67          & 89.25          & 90.50          \\
		SRC     & 56.67          & 71.83          & 75.67          & 77.92          & 82.25          & 84.00          \\
		LRC     & 57.50          & 68.08          & 70.50          & 71.75          & 73.67          & 73.92          \\
		NN      & 45.17          & 51.00          & 53.17          & 54.58          & 56.92          & 57.17          \\
		\hline
	\end{tabular}
\end{table}

\begin{table}[]
	\caption{Recognition accuracy (\%) of different methods on the AR database with the occlusion of scarves.}
	\label{tab-2}
	\centering
	\begin{tabular}{ccccccc}
		\hline
		Dim & 25& 50& 75& 100& 200& 300\\
		\hline
		DLRR-CR & \textbf{58.25} & \textbf{84.75} & \textbf{88.50} & \textbf{90.83} & \textbf{91.58} & \textbf{91.83} \\
		LRR-CR  & 53.50          & 82.58          & 87.75          & 89.25          & 89.67          & 89.50          \\
		LRR-CRC & 46.08          & 76.17          & 83.33          & 86.25          & 90.67          & 90.75          \\
		CRC     & 45.08          & 72.25          & 80.50          & 84.75          & 90.00          & 90.33          \\
		SRC     & 51.42          & 66.25          & 70.75          & 74.75          & 79.17          & 80.58          \\
		LRC     & 56.42          & 65.67          & 68.08          & 70.00          & 70.58          & 70.50          \\
		NN      & 39.75          & 45.42          & 47.00          & 48.83          & 50.50          & 50.75          \\
		\hline
	\end{tabular}
\end{table}

\begin{table}[]
	\caption{Recognition accuracy (\%) of different methods on the AR database with the occlusion of sunglasses and scarves.}
	\label{tab-3}
	\centering
	\begin{tabular}{ccccccc}
		\hline
		Dim & 25& 50& 75& 100& 200& 300\\
		\hline
		DLRR-CR & \textbf{55.82} & \textbf{81.53} & \textbf{87.06} & \textbf{88.59} & \textbf{90.65} & \textbf{90.29} \\
		LRR-CR  & 53.82          & 78.29          & 85.53          & 88.00          & 88.82          & 88.53          \\
		LRR-CRC & 45.65          & 72.76          & 80.29          & 85.53          & 89.71          & 90.12          \\
		CRC     & 42.94          & 69.29          & 78.35          & 82.12          & 88.18          & 89.53          \\
		SRC     & 51.06          & 66.00          & 70.88          & 73.65          & 78.06          & 80.47          \\
		LRC     & 53.53          & 64.71          & 68.35          & 69.41          & 70.94          & 70.76          \\
		NN      & 35.47          & 40.65          & 42.65          & 44.12          & 46.41          & 47.06          \\
		\hline
	\end{tabular}
\end{table}

\subsection{Extended Yale B database}
\label{4.2}
The Extended Yale B database \cite{naseem2010linear} includes 2414 frontal face images for 38 subjects, each subject has 64 face images obtained under different laboratory-controlled lighting conditions. The face images are cropped to a size of 192$\times$168 pixels and normalized in advance. Some exampe face images from the the extended Yale B database are shown in Fig.~\ref{fig:exam_img} (b). Firstly, 16 face images are randomly selected from each individual for training, and the remaining images for testing. Secondly, 32 face images are randomly selected from each subject for training, and the remaining images for testing. The eigenface feature dimensions are set to the same as in the experiments in the AR database. The regularization parameters used in DLRR-CR are set as $\beta$=1.1, $\eta$=0.001. Depending on feature dimension, the parameter $\lambda$ ranges from 0.004 to 0.1 for the two cases. All experiments run 5 times and the averaged accuracy is reported for performance evaluation shown in Tables~\ref{tab-4}-\ref{tab-5}.

From the results in Tables~\ref{tab-4}-\ref{tab-5}, for each dimension, DLRR-CR outperforms the other methods, this indictes our method can handle the problem of changes in illumination and expression. It should be noted that in the step 5 of Algorithm~\ref{alg2}, the original training samples are used to reduce dimensions. The main reason is that we have already learned a desirable PCA subspace by the derived clean dictionary with discrimination and this subspace would not be too sensitive to sparse errors. The second reason is that CRC represents test sample collaboratively by all classes, a small proportion of corrupted training samples will have a small influence and there are also abundant images taken under well controlled settings which can participate in representing the test sample.

\begin{table}[]
	\caption{Recognition accuracy (\%) of different methods on the Extended Yale B database with 16 training images per person.}
	\label{tab-4}
	\centering
	\begin{tabular}{ccccccc}
		\hline
		Dim & 25& 50& 75& 100& 200& 300\\
		\hline
		DLRR-CR & \textbf{83.61} & \textbf{92.43} & \textbf{94.57} & \textbf{95.04} & \textbf{97.19} & \textbf{97.82} \\
		LRR-CR  & 72.43          & 91.20          & 93.36          & 94.68          & 94.74          & 95.85          \\
		LRR-CRC & 63.16          & 81.77          & 88.19          & 91.31          & 95.16          & 96.27          \\
		CRC     & 61.17          & 79.20          & 85.60          & 89.17          & 94.21          & 96.09          \\
		SRC     & 45.16          & 67.00          & 77.95          & 84.29          & 93.36          & 95.07          \\
		NN      & 32.52          & 42.36          & 46.16          & 48.43          & 51.900         & 52.73          \\
		\hline
	\end{tabular}
\end{table}

\begin{table}[]
	\caption{Recognition accuracy (\%) of different methods on the Extended Yale B database with 32 training images per person.}
	\label{tab-5}
	\centering
	\begin{tabular}{ccccccc}
		\hline
		Dim & 25& 50& 75& 100& 200& 300\\
		\hline
		DLRR-CR & \textbf{89.26} & \textbf{97.83} & \textbf{98.22} & \textbf{98.86} & \textbf{99.39} & \textbf{99.61} \\
		LRR-CR  & 75.88          & 96.27          & 96.94          & 96.94          & 98.38          & 98.99          \\
		LRR-CRC & 60.18          & 84.03          & 90.07          & 93.21          & 96.30          & 97.11          \\
		CRC     & 65.44          & 85.23          & 90.62          & 93.66          & 97.02          & 97.97          \\
		SRC     & 48.05          & 72.73          & 82.08          & 87.31          & 94.99          & 96.61          \\
		NN      & 39.93          & 52.03          & 56.93          & 59.85          & 64.38          & 65.19          \\
		\hline
	\end{tabular}
\end{table}

\subsection{FR with artificial corruption}
\label{sec-4.3}
\begin{figure}[h]
\centering
\includegraphics[trim={0mm 0mm 0mm 0mm},clip, width = .8\textwidth]{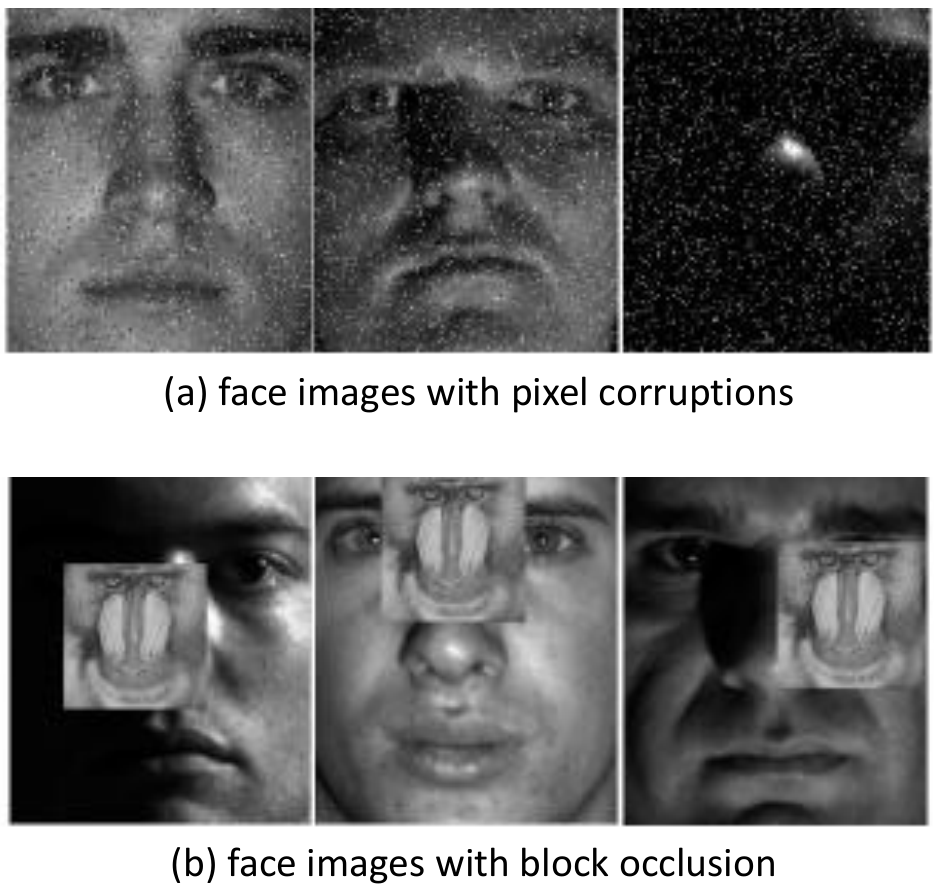}
\caption{Example images with artificial corruption.}
\label{fig:corruption}
\end{figure}

In this subsection, we consider the scenario in which both the training and test images are corrupted due to artificial occlusion. The face images from the Extended Yale B database are used to investigate the robustness of all competing approaches. In the first kind of setting, we randomly select 10\% of all the images, then we randomly select pixels of these images, and these pixels are replaced by a random value in the range of [0,1]. In the second kind of setting, to examine the robustness of block occlusion, we also randomly selected 10\% of all the samples from the database, then we randomly select square blocks of these images, and these square blocks are replaced by an unrelated image. Some representative examples of images with these two kinds of artificial occlusion are shown in Fig.~\ref{fig:corruption}. The percentage of corrupted samples in both situation are set to 10\% and 20\%.

When the artificial occlusion is added, 32 images are randomly selected from each class for training and the rest for testing. Now the regularization parameters are set as $\beta$=1.1,  $\eta$=0.001 and depend on different feature dimensions, $\lambda$ ranges from 0.005 to 0.1. We investigate the classification accuracy with the eigenface feature dimensions set as 50, 100 and 300. We run 5 times of all experiments and the average accuracy is recorded. Table~\ref{table:tab-6} shows the recognition accuracy of all seven algorithms for the two kinds of percentages of pixel corruptions. Specifically, our method achieves the best recognition performance, which are higher than those of the other methods. The recognition accuracy of block occlusion is shown in Table~\ref{table:tab-7}. We can also see that the performance gains of our algorithm is significant.

\begin{table}[]
\caption{Recognition accuracy (\%) of different methods on the Extended Yale B database with pixel corruptions.}
\label{table:tab-6}
\centering
\begin{tabular}{ccccccc}
\hline
\multirow{2}{*}{Dim} & \multicolumn{3}{c}{10\%} & \multicolumn{3}{c}{20\%} \\ \cline{2-7} 
                     & 50     & 100    & 300    & 50     & 100    & 300    \\ \hline
DLRR-CR & \textbf{96.69} & \textbf{98.28} & \textbf{99.08} & \textbf{95.10} & \textbf{96.99} & \textbf{98.36} \\
DLRR-SRC     & 94.38          & 96.88          & 97.94          & 92.85          & 95.94          & 95.88      \\
LRR-CRC    & 88.84          & 95.66          & 98.55          & 87.76          & 94.88          & 97.94       \\
CRC            & 86.95          & 93.10          & 96.85          & 85.62          & 93.04          & 95.27     \\
SRC   & 72.76          & 86.70          & 94.57          & 70.31          & 85.00          & 90.93       \\
LRC  & 92.21          & 93.52          & 94.27          & 91.37          & 93.10          & 93.18    \\
NN  & 51.45          & 58.51          & 62.99          & 51.45          & 57.09          & 57.98         \\ \hline
\end{tabular}
\end{table}

\begin{table}[]
\caption{Recognition accuracy (\%) of different methods on the Extended Yale B database with block occlusions.}
\label{table:tab-7}
\centering
\begin{tabular}{ccccccc}
\hline
\multirow{2}{*}{Dim} & \multicolumn{3}{c}{10\%} & \multicolumn{3}{c}{20\%} \\ \cline{2-7} 
                     & 50     & 100    & 300    & 50     & 100    & 300    \\ \hline
DLRR-CR & \textbf{95.83} & \textbf{97.74} & \textbf{98.78} & \textbf{94.46} & \textbf{96.55} & \textbf{97.30} \\
DLRR-SRC     & 93.57          & 96.33          & 97.80          & 91.93          & 95.13          & 96.46      \\
LRR-CRC    & 84.73          & 92.85          & 97.36          & 82.39          & 90.82          & 95.88       \\
CRC           & 81.58          & 91.62          & 96.72          & 79.16          & 89.18          & 94.46     \\
SRC    & 67.75          & 84.20          & 95.60          & 65.38          & 82.41          & 93.35       \\
LRC  & 86.06          & 89.96          & 92.07          & 82.33          & 85.81          & 88.34    \\
NN  & 49.78          & 57.71          & 62.88          & 46.38          & 54.20          & 59.10         \\ \hline
\end{tabular}
\end{table}

\section{Conclusions}
\label{sec-5}
Collaborative representation mechanism applied in face recognition has aroused considerable interest during the past few years. The most challenging case of robust face recognition is both the training and test images are corrupted by the unknown type of corruptions. A discriminative low-rank representation method for collaborative representation-based (DLRR-CR) method is proposed to solve this challenging problem. Our key contribution is the use of structural incoherence term which promotes the discrimination between different subjects. Meanwhile, CRC can achieve superb performance in various pattern classification tasks with low computational complexity. The experimental results proved that DLRR-CR is robust and effective to the possible corruptions existed in the face images.

\section*{Acknowledgments}
This work was supported by the National Natural Science Foundation of China (Grant No. 61672265, U1836218), the 111 Project of Ministry of Education of China (Grant No. B12018), the Postgraduate Research \& Practice Innovation Program of Jiangsu Province under Grant No. KYLX\_1123, the Overseas Studies Program for Postgraduates of Jiangnan University and the China Scholarship Council (CSC, No.201706790096).

\section*{References}

\bibliography{mybibfile}

\end{document}